\pdfoutput=1
\documentclass[conference]{IEEEtran}

\usepackage{amsmath,amssymb}
\usepackage{booktabs}
\usepackage{graphicx}
\usepackage{url}
\usepackage[hidelinks]{hyperref}
\usepackage{array}
\usepackage{tikz}
\usetikzlibrary{arrows.meta}

\newcommand{\method}[1]{\texttt{#1}}

\title{Matched-Input Estimates Differ in Sign Across Architectures:\\
Auditing EEG Foundation Models on Motor Imagery}

\author{%
\IEEEauthorblockN{Kevin Zhou}
\IEEEauthorblockA{Hopewell Valley Central High School\\
\textit{High School Student Author}\\
zhouk.2009@gmail.com}
\and
\IEEEauthorblockN{Sparsh Roy}
\IEEEauthorblockA{Hopewell Valley Central High School\\
\textit{High School Student Author}\\
sparshr@mit.edu}}

\begin{document}
\maketitle

\begin{abstract}
Pretrained EEG foundation models are increasingly proposed as general-purpose encoders for brain--computer interfaces, yet recent benchmarks disagree about when their representations transfer to downstream tasks. We audit LaBraM and CBraMod on motor imagery under a validation-locked protocol in which preprocessing, architecture, optimization, freeze depth, checkpoint, temperature, and method selection are determined using training-session data only. On four-class BCI Competition IV-2a, every supervised comparator evaluated here outperforms every foundation-model configuration, including validation-selected fine-tuning.

We then examine a key confound: foundation models and task-specific decoders are normally evaluated using different input pipelines. Retraining three supervised architectures on the broadband arrays consumed by the foundation models produces matched-input accuracy differences of opposite sign across architectures: broadband input improves ATCNet by 0.078 accuracy while reducing EEG Conformer accuracy by 0.088. None of the three individual matched-input terms is significant after multiple-comparison correction at $n=9$, so we treat the sign variation descriptively rather than as a formal architecture-by-pipeline interaction. These observed sign differences suggest that a single comparator may not provide an architecture-invariant decomposition of a pretrained-versus-supervised performance gap.

The four-class deficit also does not reproduce uniformly across motor-imagery datasets: on two-class BNCI2014-004 we cannot detect the same separation between fine-tuned CBraMod and the supervised comparators. Finally, validation-fitted temperature scaling returns foundation-model calibration error to the supervised range despite substantially lower four-class accuracy.
\end{abstract}


\section{Introduction}

Large pretrained EEG encoders such as LaBraM \cite{jiang2024labram} and CBraMod \cite{wang2025cbramod} are intended to provide reusable representations across downstream neural-decoding tasks. Motor imagery is an informative stress test because subject-specific datasets are small, session shift is substantial, and strong supervised decoders already exist \cite{pfurtscheller1999erd,lotte2018review}. Recent EEG foundation-model benchmarks, however, do not give a consistent answer about transfer \cite{kontras2026neuroatlas,zare2026negativecontrol}.

Two methodological problems complicate comparison. First, evaluation protocols differ. Some published BCI Competition IV-2a results use protocols that differ from the official cross-session setting---for example within-session random splits or other choices that make headline accuracies difficult to compare directly \cite{yu2022lffn}. Standardized BCI benchmarking efforts were developed in part to address this broader reproducibility problem \cite{jayaram2018moabb,chevallier2024reproducibility}.

Second, pretrained and task-specific models often require different input pipelines. A foundation model may expect broadband recordings at a particular sampling rate and scale, whereas a supervised motor-imagery decoder may be optimized for a narrower frequency band. A direct performance difference therefore combines model and input-pipeline effects. NeuroAtlas explicitly identifies EEG-specific preprocessing as a possible confound \cite{kontras2026neuroatlas}, but the corresponding term is rarely examined across several comparator architectures.

We address both issues. Every tunable decision is selected using training-session data only, with machine-checked provenance preventing evaluation labels from entering selection (Fig.~\ref{fig:protocol}). We then repeat the matched-input control for three supervised architectures rather than one. Our contributions are: (i) a validation-locked evaluation protocol whose leakage guard is encoded in the stored artifacts; (ii) an audit of LaBraM and CBraMod against three supervised motor-imagery architectures plus an architecture-matched random-initialization control; (iii) a three-architecture matched-input analysis whose point estimates differ in sign across comparators; (iv) evidence that the large BCI IV-2a deficit does not reproduce uniformly on BNCI2014-004, without attributing the difference to class count alone; and (v) calibration analysis showing that a single validation-fitted temperature can return foundation-model ECE to the supervised range even when accuracy remains substantially lower.

\begin{figure}[t]
\centering
\resizebox{\columnwidth}{!}{%
\begin{tikzpicture}[
    >=Latex,
    node font=\scriptsize,
    box/.style={draw,rounded corners=1.5pt,align=center,minimum height=6mm,inner xsep=4pt,inner ysep=2.5pt},
    arr/.style={->,line width=.45pt}
]
\node[font=\footnotesize\bfseries,anchor=west] at (0,2.25) {(a) Validation-locked evaluation};
\node[box] (train) at (0.6,1.35) {Training\\session};
\node[box] (fit)   at (2.65,1.75) {Fit split\\80\%};
\node[box] (sel)   at (2.65,0.95) {Selection split\\20\%};
\node[box,text width=2.3cm] (lock) at (5.2,1.35) {Lock configuration\\architecture, LR,\\checkpoint, temperature};
\node[box] (test) at (7.75,1.35) {Held-out\\session};
\node[box] (report) at (9.65,1.35) {Report\\metrics once};
\draw[arr] (train.east) -- (fit.west);
\draw[arr] (train.east) -- (sel.west);
\draw[arr] (fit.east) -- (lock.west);
\draw[arr] (sel.east) -- (lock.west);
\draw[arr] (lock.east) -- node[above,font=\tiny] {configuration frozen} (test.west);
\draw[arr] (test.east) -- (report.west);
\node[font=\tiny,align=center] at (6.5,0.55) {held-out labels are never used for selection};

\node[font=\footnotesize\bfseries,anchor=west] at (0,-0.15) {(b) Matched-input control};
\node[box] (eeg) at (0.6,-1.05) {Same EEG\\recordings};
\node[box] (narrow) at (2.95,-0.65) {Narrowband\\8--30 Hz, 250 Hz};
\node[box] (broad) at (2.95,-1.45) {Broadband\\0.1--75 Hz, 200 Hz};
\node[box,text width=2.15cm] (arch) at (5.65,-1.05) {Same supervised\\architecture\\independently tuned};
\node[box,text width=2.0cm] (term) at (8.15,-1.05) {Matched-input term\\narrow $-$ broad};
\draw[arr] (eeg.east) -- (narrow.west);
\draw[arr] (eeg.east) -- (broad.west);
\draw[arr] (narrow.east) -- (arch.west);
\draw[arr] (broad.east) -- (arch.west);
\draw[arr] (arch.east) -- (term.west);
\node[font=\tiny,align=center] at (5.65,-2.05) {Repeated for ShallowConvNet, ATCNet, and EEG Conformer};
\end{tikzpicture}%
}
\caption{Experimental protocol. (a) All tunable decisions are selected from the training session; the held-out session is evaluated only after the configuration is locked. (b) Each supervised comparator is trained on narrowband and broadband pathways; the matched-input term is narrowband minus broadband accuracy.}
\label{fig:protocol}
\end{figure}

\section{Related Work}

\textbf{EEG foundation models.}
LaBraM \cite{jiang2024labram} and CBraMod \cite{wang2025cbramod} are transformer encoders pretrained on large collections of unlabeled EEG. Recent benchmarks include AdaBrain-Bench \cite{wu2025adabrainbench}, EEG-FM-Bench \cite{xiong2025eegfmbench}, Brain4FMs \cite{shen2026brain4fms}, NeuralBench \cite{banville2026neuralbench}, NeuroAtlas \cite{kontras2026neuroatlas}, OmniEEG-Bench \cite{lu2026omnieegbench}, and Zare's negative-control analysis \cite{zare2026negativecontrol}. Their conclusions are not uniform. In particular, recent audits already use matched random-weight encoders \cite{kontras2026neuroatlas,banville2026neuralbench,zare2026negativecontrol}; we adopt that control rather than claim it as new.

\textbf{Input pipelines as a confound.}
NeuroAtlas notes that EEG-specific preprocessing may affect reported results \cite{kontras2026neuroatlas}. We measure the corresponding input-matching difference by retraining each supervised comparator on the broadband arrays supplied to the foundation models. Because the pathways differ in band, sampling rate, filter design, notching, scaling, and validation-selected retuning, we call the quantity the \emph{matched-input term}, not an isolated preprocessing effect.

\textbf{Supervised motor-imagery decoders.}
Classical CSP and filter-bank CSP remain established references \cite{ramoser2000csp,ang2008fbcsp}, while EEGNet \cite{lawhern2018eegnet} is a widely used compact neural baseline. Our evaluated comparator set is ShallowConvNet \cite{schirrmeister2017deep}, ATCNet \cite{altaheri2023atcnet}, and EEG Conformer \cite{song2023conformer}.

\section{Validation-Locked Protocol}

\subsection{Selection and Enforcement}

One rule is applied throughout: every tunable choice---preprocessing parameters, standardization statistics, architecture, probe regularization, learning rate, freeze depth, stopping epoch, temperature, and downstream thresholds---is determined using data from the training session only. Held-out-session labels are used only to compute final metrics.

Every retained result row contains an \method{eval\_labels\_used\_for\_selection} field written by the runner. The table-building code refuses artifacts for which that field is missing or true, and probability files are rechecked before calibration or selective-prediction metrics are computed. The leakage constraint is therefore encoded in the artifact format rather than relying only on author discipline.

\subsection{Splits, Inputs, and Provenance}

For BCI IV-2a, eight of nine subjects contribute 230 fit, 58 selection, and 288 held-out-session trials; subject 9 contributes 221, 56, and 264 after artifact rejection. The small selection split is relevant to the fine-tuning instability observed below.

The standard supervised pathway uses 8--30 Hz EEG at 250 Hz, standardized per channel using fit-split statistics. The foundation-model pathway uses 0.1--75 Hz EEG resampled to 200 Hz with a 50 Hz notch and microvolt scaling. Each foundation model therefore receives the input type for which its pretrained encoder was designed.

Foundation models also use learned electrode representations, so channel identity is part of the input contract. The canonical montage is reconstructed and checked against all source files. Each run records device, software versions, and determinism flags; all values reported here come from the deterministic CPU backend, preventing results from different execution backends from being silently pooled.

\section{Experimental Design}

\subsection{Models and Adaptation}

We evaluate LaBraM-base (5.8M parameters) \cite{jiang2024labram} and CBraMod (5.0M) \cite{wang2025cbramod}. LaBraM is ported from the reference release; CBraMod uses the community Braindecode implementation, reducing dependence on a local reimplementation. Two additional implementations were attempted but are treated as scope rather than negative evidence: Braindecode LaBraM rejects our input length because of its temporal-embedding configuration, and EEGPT's public interface did not expose the interpolation target needed for our input format.

Supervised comparator families are ShallowConvNet \cite{schirrmeister2017deep}, ATCNet \cite{altaheri2023atcnet}, and EEG Conformer \cite{song2023conformer}, each tuned on the selection split. For the foundation models we evaluate frozen probing and full/partial fine-tuning. Fine-tuning selects learning rate, freeze depth, warm-up behavior, and stopping epoch on validation data, with layer-wise decay 0.9, weight decay 0.05, label smoothing 0.1, batch size 16, and a 60-epoch maximum. Validation favors ten frozen LaBraM blocks at $10^{-4}$ but full CBraMod unfreezing at $5\times10^{-4}$, illustrating that ``foundation-model fine-tuning'' is not one transferable recipe.

\subsection{Controls}

\textbf{Architecture-matched random initialization.}
The same encoder architecture is trained from random weights, separating two explanations for poor downstream performance: weak task-relevant information in the pretrained weights versus an architecture that is intrinsically difficult to learn from a few hundred labeled trials. We verify checkpoint loading: 178 of 179 tensors differ from a fresh initialization.

\textbf{Matched input.}
ShallowConvNet, ATCNet, and EEG Conformer are each retrained on the broadband arrays consumed by the foundation models. Temporal kernel and filter count are selected per subject on validation data. Because tuning occurs independently for each pathway, the matched-input term captures the input pipeline together with the architecture retuning required by that pipeline; it should not be read as the isolated effect of any single filter or preprocessing operation.

\subsection{Datasets and Statistics}

BCI Competition IV-2a is the primary dataset \cite{brunner2008graz,tangermann2012competition}: nine subjects, two sessions, four motor-imagery classes, and 22 EEG channels. BNCI2014-004 \cite{leeb2007bnci004} provides an independent second setting with nine subjects, two classes, and three electrodes (C3, Cz, C4). BNCI2014-001 is not used as an independent replication because it reconstructs the same underlying recordings as BCI IV-2a. BNCI2014-004 is not a controlled class-count ablation because dataset identity and channel count change simultaneously.

Brier score is the primary probability-quality metric; we additionally report accuracy, ECE, and negative log-likelihood \cite{brier1950verification,gneiting2007proper}. Temperature scaling \cite{guo2017calibration} is fit on validation data only and capped at 20.0; saturation at the cap is recorded because it indicates severe raw-logit overconfidence. Pairwise comparisons use two-sided subject-paired Wilcoxon signed-rank tests \cite{wilcoxon1945individual} with Benjamini--Hochberg correction \cite{benjamini1995fdr} within each reported primary family. At $n=9$, the smallest attainable two-sided exact Wilcoxon value is $2^{1-n}=0.0039$. We distinguish failure to detect a difference from evidence of equality throughout.

\section{Results}

\subsection{Four-Class Motor Imagery}

Table~\ref{tab:main} summarizes representative configurations from a 26-configuration validation-locked grid. Every supervised comparator exceeds every foundation-model configuration in accuracy. Validation-selected CBraMod fine-tuning reaches 0.4796, compared with 0.5305 for broadband EEG Conformer and 0.6962 for ShallowConvNet. Against each supervised comparator, the paired accuracy comparison with the best foundation-model configuration reaches the attainable exact-Wilcoxon floor, with no subject-level wins for the foundation-model arm.

\begin{table}[t]
\centering
\caption{Held-out BCI IV-2a performance ($n=9$; chance accuracy 0.25). Selected configurations from the full grid.}
\label{tab:main}
\small
\begin{tabular}{lccc}
\toprule
Configuration & Acc. & Brier & ECE \\
\midrule
ShallowConvNet           & 0.6962 & 0.4047 & 0.0808 \\
ATCNet, broadband        & 0.6562 & 0.4695 & 0.1062 \\
EEG Conformer, broadband & 0.5305 & 0.5858 & 0.1072 \\
CBraMod, fine-tuned      & 0.4796 & 0.6493 & 0.1236 \\
CBraMod, frozen          & 0.3927 & 0.7080 & 0.1030 \\
CBraMod, random init.    & 0.3407 & 0.7396 & 0.0993 \\
LaBraM, frozen           & 0.3322 & 0.7323 & 0.0874 \\
\bottomrule
\end{tabular}
\end{table}

\subsection{Fine-Tuning Stability}

Under the initial reference recipe, the validation-optimal epoch occurs between epochs 0 and 4 across runs even though warm-up lasts six epochs: the model peaks before reaching its nominal learning rate. A six-configuration search ranked entirely on validation data improves fine-tuning substantially without reversing the ordering in Table~\ref{tab:main}. The successful changes---freezing additional blocks and lowering learning rate---both limit displacement from pretrained weights, consistent with overfitting under a roughly 230-trial subject-specific fit budget.

\subsection{Matched-Input Estimates Differ in Sign}

We define the matched-input term as narrowband minus broadband accuracy for the same comparator family. A negative value therefore means the broadband foundation-model pathway performs better. Table~\ref{tab:matched} reports the three comparators.

\begin{table}[t]
\centering
\caption{Matched-input analysis on BCI IV-2a ($n=9$). Wins count subjects for which narrowband input achieved higher accuracy.}
\label{tab:matched}
\small
\begin{tabular}{lccccc}
\toprule
Architecture & Narrow & Broad & Diff. & Wins & $p_{\mathrm{BH}}$ \\
\midrule
ATCNet         & .5788 & .6562 & $-.078$ & 1/9 & .073 \\
ShallowConvNet & .6453 & .6084 & $+.037$ & 5/9 & .304 \\
EEG Conformer  & .6182 & .5305 & $+.088$ & 8/9 & .176 \\
\bottomrule
\end{tabular}
\end{table}

For ATCNet, broadband input improves mean accuracy by 0.078 and wins for eight of nine subjects. For EEG Conformer, the direction reverses: narrowband input improves mean accuracy by 0.088 and wins for eight of nine subjects. ShallowConvNet lies between these cases. None of the three individual matched-input terms survives Benjamini--Hochberg correction ($p_{\mathrm{BH}}=0.073$, $0.304$, and $0.176$, respectively), so we do not claim a formal architecture-by-input interaction. Although this does not establish an interaction, the observed matched-input estimate changes sign across comparator architectures. A single-architecture control should therefore be interpreted cautiously when used to apportion a pretrained-versus-supervised performance gap.

\subsection{Task Dependence and Robustness}

On BNCI2014-004, fine-tuned CBraMod reaches 0.8085 accuracy, compared with 0.8458 for the best supervised arm, 0.8185 for tuned ShallowConvNet, and 0.7981 for EEG Conformer. We therefore cannot detect the same uniform foundation-model deficit observed on BCI IV-2a. This difference is not attributed to class count alone because the datasets also differ in recordings and electrode count.

The decisive fine-tuned-CBraMod versus random-initialization comparison was repeated under three independent full-pipeline reseeds, redrawing the fit/selection split, initialization, and batch order. Subject-averaged effects are 0.139, 0.132, and 0.130 on BCI IV-2a and 0.107, 0.111, and 0.108 on BNCI2014-004. Each of the six paired comparisons is individually significant under the same subject-level test, and the lowest subject win count is eight of nine. These are robustness checks rather than a separate corrected confirmatory family; seed-to-seed spread is not treated as inferential uncertainty across subjects.

To test whether the random-initialization arm was simply under-trained, we repeated the BCI IV-2a comparison for 240 epochs for both arms. Random initialization gained 0.017 accuracy and closed 0.024 of the gap, while the pretrained arm did not improve. Random initialization reached its best validation Brier near epoch 218 and was approximately flat from epoch 190 onward; its best value, 0.755, remained near the four-class chance Brier of 0.750.

\subsection{Calibration}

Raw foundation-model logits are strongly overconfident. Frozen configurations require validation-fitted temperatures from 3.5 to 9.5, with one to four of nine subjects reaching the 20.0 cap, compared with approximately 1.03 for a tuned supervised convolutional model. After temperature scaling, held-out CBraMod ECE falls between 0.103 and 0.124, within the 0.101--0.143 range observed across tuned supervised comparators despite much lower four-class accuracy.

\section{Discussion and Limitations}

The central methodological observation is not that one input pipeline is universally better. In this sample, the matched-input point estimate ranges from a 0.078 improvement for ATCNet to a 0.088 reduction for EEG Conformer. None of those individual terms survives multiple-comparison correction, and we do not claim a formal architecture-by-pipeline interaction. Their opposite observed directions nevertheless show why a decomposition based on one comparator should be interpreted cautiously. A practical remedy is inexpensive: repeat matched-input analysis across multiple reasonable comparator architectures and report raw differences rather than a percentage ``pipeline share,'' whose interpretation becomes unstable when the underlying term changes sign.

The main performance result is similarly bounded. These experiments do not show that pretrained EEG encoders fail generally. They show that, under a leakage-controlled cross-session protocol, the two encoders evaluated here trail three supervised architectures on four-class BCI IV-2a, while fine-tuned CBraMod is competitive with the supervised arms on BNCI2014-004. Calibration also emerges as a separate axis: one validation-fitted scalar largely repairs probability calibration even when discrimination remains weak. Whether that correction remains stable under reduced-precision on-device inference is untested here.

Several limitations matter. We successfully evaluate two foundation models out of four attempted implementations and only two motor-imagery datasets. BNCI2014-004 contains three electrodes and is not a controlled replication of the 22-channel BCI IV-2a setting. The decisive pretrained-versus-random comparison uses three full reseeds, whereas several other configurations use one seed. Subject-level inference is limited to nine participants, making the design better suited to large, consistent effects than small improvements. The matched-input analysis re-tunes two architectural hyperparameters per pathway, so its difference captures input pipeline plus validation-selected retuning rather than an isolated preprocessing operation; the broadband search also does not exhaust the comparator architecture space. Finally, we do not perform a formal architecture-by-pipeline interaction test, so the cross-architecture sign difference is reported as a descriptive robustness concern rather than a confirmed interaction.

\section*{Reproducibility}

Code, per-subject result artifacts, per-trial probability files, pinned library versions, and a manifest with SHA256 hashes accompany the study. Every reported table is regenerated directly from those artifacts by analysis scripts; no table value is transcribed manually.
\section*{AI Use Disclosure}

The authors used large language models for language editing after the
study was completed and the manuscript drafted. Its use was limited to
wording and readability in the prose sections. It contributed no study
design, data, analysis, results, figures, tables, or code, and was not
applied to the reference list. The authors reviewed all edited text and
are responsible for the content of this paper.

\bibliographystyle{IEEEtran}
\bibliography{refs}

\end{document}